\documentclass[conference]{IEEEtran}
\IEEEoverridecommandlockouts
\usepackage{cite}
\usepackage{amsmath,amssymb,amsfonts}
\usepackage{amsthm}
\usepackage[english]{babel}

\usepackage{algorithmic}
\usepackage{graphicx}
\usepackage{subcaption}
\usepackage{textcomp}
\usepackage{xcolor}
\usepackage{hyperref}
\def\BibTeX{{\rm B\kern-.05em{\sc i\kern-.025em b}\kern-.08em
    T\kern-.1667em\lower.7ex\hbox{E}\kern-.125emX}}
\begin{document}

\title{Laplacian-LoRA: Delaying Oversmoothing in Deep GCNs via Spectral Low-Rank Adaptation \\
}

\author{\IEEEauthorblockN{1\textsuperscript{st} Sai Vamsi Alisetti}
\IEEEauthorblockA{\textit{Department of Computer Science} \\
\textit{University of California, Santa Barbara}\\
Santa Barbara, United States \\
saivamsi@ucsb.edu}
}

\maketitle

\begin{abstract}
Oversmoothing is a fundamental limitation of deep graph convolutional networks (GCNs), causing node representations to collapse as depth increases. While many prior approaches mitigate this effect through architectural modifications or residual mechanisms, the underlying spectral cause of oversmoothing is often left implicit.

We propose \emph{Laplacian-LoRA}, a simple and interpretable low-rank spectral adaptation of standard GCNs. Rather than redesigning message passing, Laplacian-LoRA introduces a learnable, spectrally anchored correction to the fixed Laplacian propagation operator, selectively weakening contraction while preserving stability and the low-pass inductive bias.

Across multiple benchmark datasets and depths, Laplacian-LoRA consistently delays the onset of oversmoothing, extending the effective depth of GCNs by up to a factor of two. Embedding variance diagnostics confirm that these gains arise from delayed representational collapse, while learned spectral analysis demonstrates that the correction is smooth, bounded, and well behaved. Our results show that oversmoothing is a depth-dependent spectral phenomenon that can be systematically delayed through modest, low-rank adaptation of the graph propagation operator.
\end{abstract}


\section{Introduction}

Graph convolutional networks (GCNs) are a foundational class of models for graph representation learning and have achieved strong performance on node classification and related tasks \cite{kipf2016semi, hamilton2017inductive}. Despite their success, GCNs suffer from a fundamental limitation of \emph{oversmoothing} i.e as depth increases, node representations become increasingly similar and eventually collapse to nearly indistinguishable embeddings \cite{li2018deeper, oono2019graph}.

While oversmoothing is often described as a problem of representations, it is more fundamentally an \emph{operator-level contraction phenomenon}. Each GCN layer applies a fixed, graph-dependent linear operator derived from the normalized Laplacian. Repeated application of this operator progressively suppresses non-constant spectral components of node features, regardless of optimization strategy or parameterization. As depth grows, this contraction dominates the learning dynamics and makes oversmoothing inevitable.

From a spectral perspective, standard GCNs repeatedly apply a low-pass filter whose magnitude is strictly less than one on all nontrivial Laplacian eigenmodes \cite{wu2019simplifying, oono2019graph}. This induces exponential decay of higher-frequency components while preserving the constant eigenmode. Although this inductive bias is beneficial at shallow depth, it becomes excessively contractive in deeper architectures, leading to rapid loss of discriminative information.

Prior work has proposed a variety of architectural strategies to mitigate oversmoothing, including residual connections \cite{li2019deepgcns}, normalization schemes \cite{rong2019dropedge}, alternative propagation operators \cite{chen2020measuring}, and rewired graph structures \cite{nt2019revisiting}. While effective in practice, these approaches typically modify the entire message-passing pipeline, making it difficult to isolate and analyze the precise mechanism by which contraction is reduced.

In this work, we take a complementary and more targeted perspective. Rather than redesigning graph convolution, we ask a simpler question:
\begin{quote}
\emph{Can oversmoothing be delayed by directly reducing the spectral contraction rate of graph propagation, while preserving its stabilizing low-frequency bias?}
\end{quote}

To address this question, we propose \textbf{Laplacian-LoRA}, a low-rank, spectrally anchored adaptation of the GCN propagation operator. Inspired by low-rank adaptation (LoRA) in large language models \cite{hu2021lora}, Laplacian-LoRA introduces a learnable low-rank correction acting directly in the Laplacian eigenspace, selectively weakening contraction on non-constant spectral modes while preserving the operator’s structure and stability. Rather than amplifying frequencies or removing smoothing, the method increases propagation eigenvalues while keeping them strictly below unity, thereby slowing though not eliminating oversmoothing. Across multiple benchmark datasets and depths, Laplacian-LoRA consistently delays performance degradation relative to standard GCNs, extending their effective depth while retaining the low-pass inductive bias.

\section{Methodology}

\subsection{Spectral View of Oversmoothing}

We begin from a spectral characterization of oversmoothing.
Let $L$ denote the normalized graph Laplacian with eigendecomposition
\[
L = U \Lambda U^\top,
\]
where $\Lambda=\mathrm{diag}(\lambda_1,\dots,\lambda_N)$.
A standard GCN layer applies the propagation operator
\[
S = I - L = U (I-\Lambda) U^\top,
\]
corresponding to the spectral filter $g_{\text{GCN}}(\lambda)=1-\lambda$.

After $t$ layers, the component of a node representation along eigenvector $u_i$ is scaled by
\[
(1-\lambda_i)^t.
\]
Since $|1-\lambda_i|<1$ for all nonzero Laplacian eigenvalues, all non-constant spectral components decay exponentially.
As depth increases, representations collapse into the low-frequency eigenspace, resulting in oversmoothing.

\subsection{Weakening Contraction Without Instability}

Delaying oversmoothing requires weakening this per-layer spectral contraction.
Consider a modified filter
\[
g(\lambda) = (1-\lambda) + \Delta(\lambda).
\]
To slow contraction while preserving stability, the modified filter must satisfy
\[
|1-\lambda| < |g(\lambda)| < 1
\quad \text{for all } \lambda \in (0,2].
\]

These constraints strongly restrict admissible corrections.
In particular, additive corrections that are not aligned with $(1-\lambda)$ either fail to uniformly weaken contraction or lead to instability near $\lambda\approx 0$.
Consequently, the only safe way to reduce contraction across the spectrum is through multiplicative scaling of the base Laplacian filter.

We therefore restrict attention to filters of the form
\[
g(\lambda) = (1-\lambda)\bigl(1+\beta(\lambda)\bigr),
\quad \beta(\lambda)\ge 0,
\]
which strictly weaken contraction while remaining stable as long as
\[
|(1-\lambda)(1+\beta(\lambda))| < 1.
\]

\subsection{Laplacian-LoRA Operator}

Laplacian-LoRA instantiates this admissible class through a spectrally anchored, low-rank residual adaptation.
Inspired by LoRA in large language models which adapts a frozen linear operator via a low-rank residual, we adapt the graph propagation operator directly in the Laplacian eigenspace.

Specifically, Laplacian-LoRA introduces a residual spectral branch with filter
\[
g_{\text{LoRA}}(\lambda;\alpha_\ell)
=
(1-\lambda)\bigl(1-\alpha_\ell\,\theta(\lambda)\bigr),
\]
where $\theta(\lambda)\in(0,1)$ is a smooth learned function.
The correction strength is \emph{depth-annealed} according to
\[
\alpha_\ell = \alpha\,\frac{\ell}{L},
\]
so that early layers remain close to standard GCN propagation while deeper layers receive stronger contraction weakening.

The GCN and Laplacian-LoRA branches are added, yielding the effective filter
\[
g_{\text{eff}}(\lambda;\alpha_\ell)
=
(1-\lambda)\bigl(2-\alpha_\ell\,\theta(\lambda)\bigr)
=
(1-\lambda)\bigl(1+\beta(\lambda)\bigr),
\]
with $\beta(\lambda)=1-\alpha_\ell\theta(\lambda)\ge 0$.

By construction,
\[
|g_{\text{eff}}(\lambda;\alpha_\ell)| \ge |g_{\text{GCN}}(\lambda)|,
\]
implying weaker per-layer contraction and delayed oversmoothing, while stability is preserved as long as
\[
\sup_{\lambda\in(0,2]} |(1-\lambda)(1+\beta(\lambda))| < 1.
\]

Since the modulation is diagonal in the Laplacian basis, the resulting operator corresponds to an additive low-rank correction in node space when restricted to leading eigenvectors.
Thus, Laplacian-LoRA can be viewed as a LoRA-style residual adaptation anchored to the graph spectrum.

\begin{figure*}[t]
    \centering

    \begin{subfigure}[t]{0.26\textwidth}
        \centering
        \includegraphics[width=\linewidth]{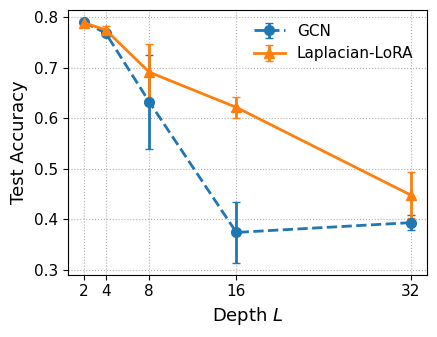}
        \caption{PubMed}
    \end{subfigure}\hfill
    \begin{subfigure}[t]{0.26\textwidth}
        \centering
        \includegraphics[width=\linewidth]{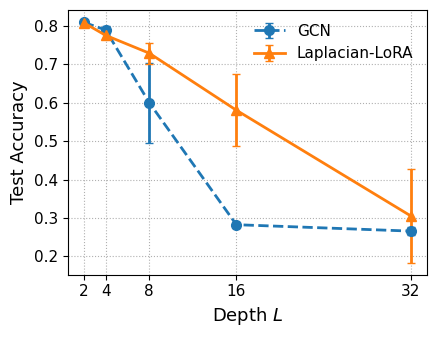}
        \caption{Cora}
    \end{subfigure}\hfill
    \begin{subfigure}[t]{0.26\textwidth}
        \centering
        \includegraphics[width=\linewidth]{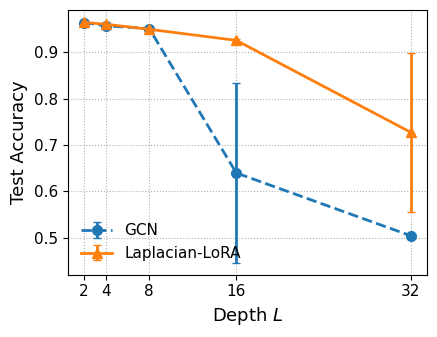}
        \caption{Coauthor Physics}
    \end{subfigure}

    \vspace{0.5em}

    \begin{subfigure}[t]{0.26\textwidth}
        \centering
        \includegraphics[width=\linewidth]{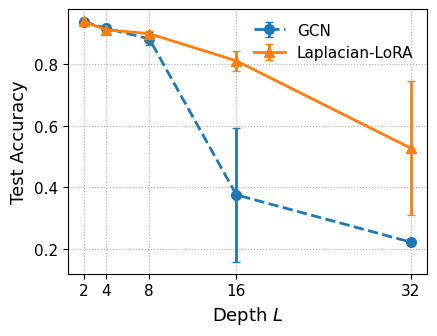}
        \caption{Coauthor CS}
    \end{subfigure}\hspace{0.04\textwidth}
    \begin{subfigure}[t]{0.26\textwidth}
        \centering
        \includegraphics[width=\linewidth]{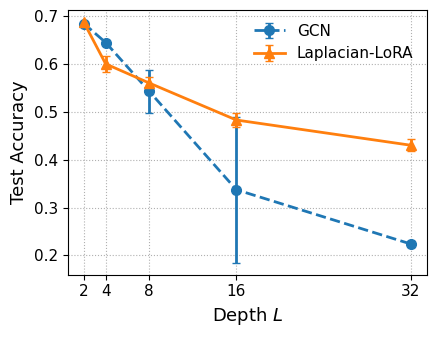}
        \caption{CiteSeer}
    \end{subfigure}

    \caption{
    Test accuracy versus depth $L$ for standard GCNs and Laplacian-LoRA across five benchmark datasets.
    Laplacian-LoRA consistently mitigates over-smoothing and depth degradation.
    }
    \label{fig:depth_vs_accuracy}
\end{figure*}

\section{Implementation Details}

We evaluate Laplacian-LoRA on five node classification benchmarks:
\textsc{Cora}, \textsc{Citeseer}, \textsc{Pubmed}, \textsc{CoauthorCS}, and \textsc{CoauthorPhysics}.
For citation networks, we use the standard Planetoid splits, while for coauthor graphs we use the official PyTorch Geometric splits. All models follow a standard GCN architecture with depths
$L \in \{2,4,8,16,32\}$, hidden dimension $64$, ReLU activations, and dropout rate $0.5$.
No batch normalization or residual connections are used in the backbone.
Models are trained with Adam (learning rate $0.01$, weight decay $5\times10^{-4}$) for up to $200$ epochs, with early stopping based on validation accuracy (patience $=50$).
Results are averaged over five random seeds. For each dataset, we precompute and cache the top $k=64$ eigenpairs of the normalized graph Laplacian using partial eigendecomposition.
Laplacian-LoRA is implemented as a shared residual branch applied at all hidden layers.
The spectral modulation function $\theta(\lambda)$ is parameterized by a two-layer MLP with hidden dimension $32$ and sigmoid output to ensure bounded corrections.
Oversmoothing is analyzed by measuring node embedding variance as a function of depth.
All experiments are implemented in PyTorch Geometric and run on a single NVIDIA A100 GPU.

\begin{figure*}[t]
    \centering

    \begin{subfigure}{0.30\textwidth}
        \centering
        \includegraphics[width=\linewidth]{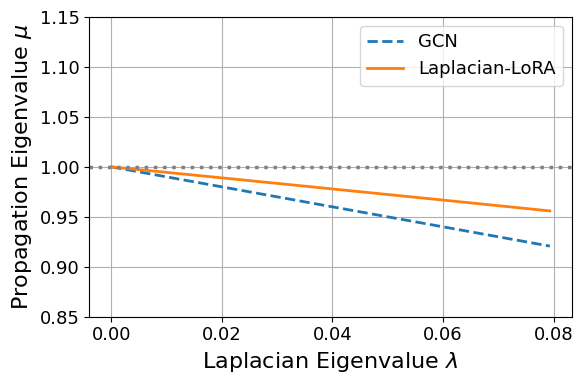}
        \caption{Propagation spectrum}
        \label{fig:prop-cora}
    \end{subfigure}
    \hfill
    \begin{subfigure}{0.30\textwidth}
        \centering
        \includegraphics[width=\linewidth]{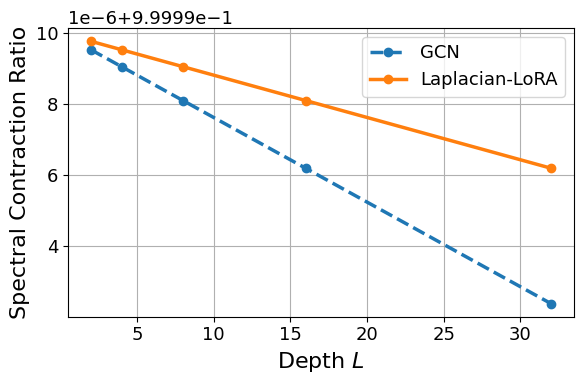}
        \caption{Spectral contraction vs.\ depth}
        \label{fig:contract-cora}
    \end{subfigure}
    \hfill
    \begin{subfigure}{0.30\textwidth}
        \centering
        \includegraphics[width=\linewidth]{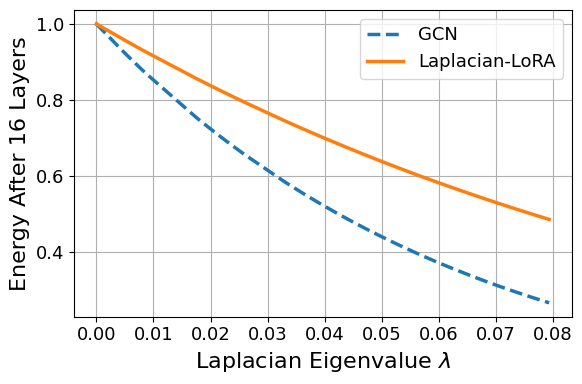}
        \caption{Energy retention ($L{=}16$)}
        \label{fig:energy-cora}
    \end{subfigure}

    \vspace{0.6em}

    \begin{subfigure}{0.30\textwidth}
        \centering
        \includegraphics[width=\linewidth]{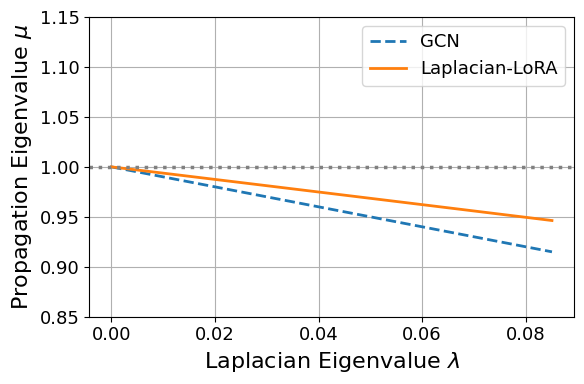}
        \caption{Propagation spectrum}
        \label{fig:prop-coauthorcs}
    \end{subfigure}
    \hfill
    \begin{subfigure}{0.30\textwidth}
        \centering
        \includegraphics[width=\linewidth]{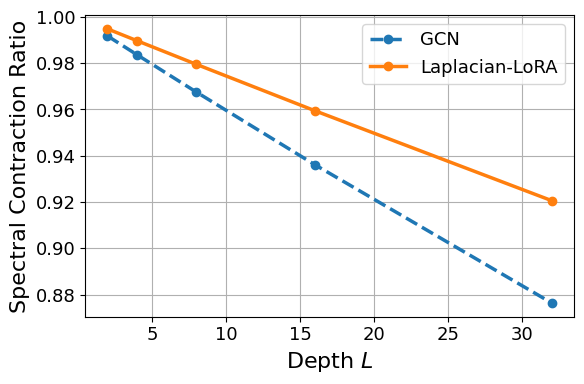}
        \caption{Spectral contraction vs.\ depth}
        \label{fig:contract-coauthorcs}
    \end{subfigure}
    \hfill
    \begin{subfigure}{0.30\textwidth}
        \centering
        \includegraphics[width=\linewidth]{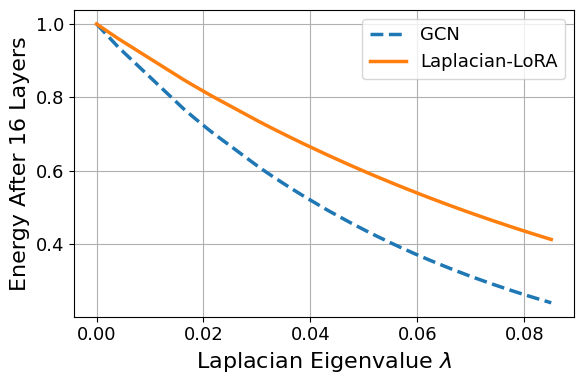}
        \caption{Energy retention ($L{=}16$)}
        \label{fig:energy-coauthorcs}
    \end{subfigure}

    \caption{
    \textbf{Spectral analysis of Laplacian-LoRA.}
    \emph{Top row (Cora):} 
    \emph{Left:} propagation eigenvalues $\mu(\lambda)$ as a function of Laplacian eigenvalue $\lambda$.
    \emph{Center:} depth-wise spectral contraction ratio $\left(|\mu_2|/|\mu_1|\right)^L$.
    \emph{Right:} retained spectral energy after $L{=}16$ layers.
    \emph{Bottom row (CoauthorCS):} corresponding spectral diagnostics on a larger graph.}
    
    \label{fig:spectral-analysis}
\end{figure*}

\section{Results}

\subsection{Accuracy as a Function of Depth}

We evaluate standard GCNs and Laplacian-LoRA across five benchmark datasets for depths ranging from $L=2$ to $L=32$ (Fig.~\ref{fig:depth_vs_accuracy}). At shallow depths ($L \leq 4$), both methods achieve comparable performance, with standard GCNs occasionally performing slightly better, consistent with the strong low-pass inductive bias of fixed Laplacian smoothing.

As depth increases, GCN accuracy degrades rapidly across all datasets, reflecting the onset of oversmoothing. In contrast, Laplacian-LoRA consistently maintains higher accuracy at intermediate depths ($L=8$-$16$), substantially extending the effective depth of graph convolution. This effect is particularly pronounced on larger and more heterogeneous graphs such as CoauthorCS and CoauthorPhysics, where GCNs exhibit sharp performance collapse while Laplacian-LoRA remains competitive.

At larger depths ($L=32$), both methods eventually degrade, indicating that Laplacian-LoRA delays but does not eliminate oversmoothing. Overall, these results demonstrate that Laplacian-LoRA significantly improves depth robustness across diverse graph domains.

\subsection{Oversmoothing Diagnostics}

To directly quantify representational collapse, we analyze the variance of node embeddings as a function of depth $L$ (Fig.~2). For standard GCNs, embedding variance decreases rapidly with depth across datasets, indicating strong contraction toward uniform representations and the onset of oversmoothing.

In contrast, Laplacian-LoRA preserves substantially higher variance at intermediate depths, demonstrating improved retention of discriminative information. Although variance eventually decreases at larger depths, this decay occurs significantly later than for standard GCNs, confirming that Laplacian-LoRA delays but does not eliminate oversmoothing.

\begin{figure}[h!]
    \centering
    \begin{subfigure}[t]{0.48\linewidth}
        \centering
        \includegraphics[width=\linewidth]{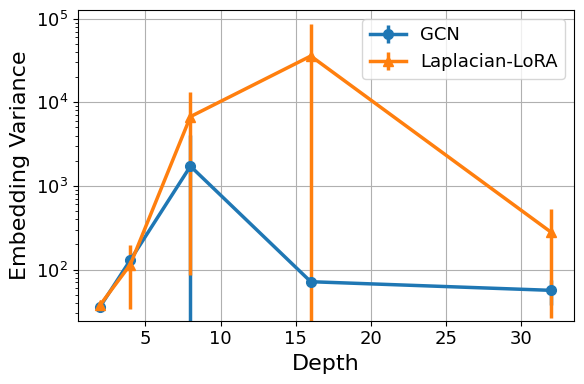}
        \caption{Cora}
        \label{fig:depth_var_cora}
    \end{subfigure}\hfill
    \begin{subfigure}[t]{0.48\linewidth}
        \centering
        \includegraphics[width=\linewidth]{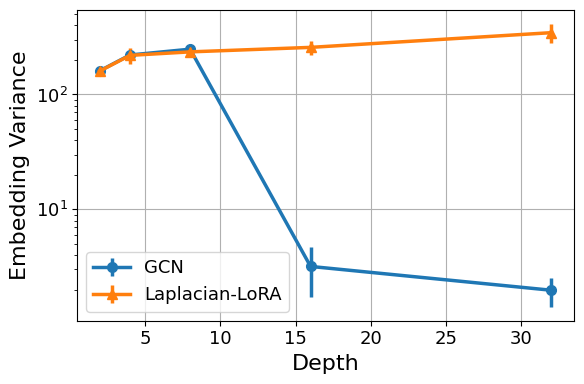}
        \caption{Coauthor CS}
        \label{fig:depth_var_coauthorcs}
    \end{subfigure}

    \caption{
    Embedding variance as a function of depth $L$ for standard GCNs and Laplacian-LoRA.}
    \label{fig:depth_vs_variance}
\end{figure}

\subsection{Learned Spectral Response}

To characterize how Laplacian-LoRA modifies message propagation, we analyze its behavior directly in the spectral domain. We examine three complementary diagnostics: 

\paragraph{Propagation Spectrum.}
A standard GCN applies a fixed low-pass spectral response, leading to propagation eigenvalues that decrease linearly with the Laplacian eigenvalue.
Laplacian-LoRA introduces a learned, spectrally aligned residual that shifts this response upward for nonzero frequencies while preserving stability.

As shown in Fig.~\ref{fig:prop-cora} and Fig.~\ref{fig:prop-coauthorcs}, the learned response remains smooth, monotonic, and bounded, and does not cross the stability boundary.
This indicates that Laplacian-LoRA weakens contraction without introducing amplification or oscillatory behavior.

\paragraph{Depth-wise Spectral Contraction.}
Oversmoothing is governed not by the spectral radius itself which remains fixed due to the constant eigenmode but by the rate at which non-constant spectral components contract relative to the dominant mode.
We quantify this effect using the depth-dependent contraction ratio
\begin{equation}
C(L)
=
\left(\frac{|\mu_2|}{|\mu_1|}\right)^L ,
\end{equation}
where $|\mu_1|$ and $|\mu_2|$ denote the largest and second-largest propagation eigenvalues in magnitude, and $L$ is the network depth.

Figures~\ref{fig:contract-cora} and \ref{fig:contract-coauthorcs} show that this ratio decays substantially faster for standard GCNs than for Laplacian-LoRA.
The slower decay induced by Laplacian-LoRA indicates delayed collapse onto the dominant eigenspace, providing a direct spectral explanation for improved depth scalability.

\paragraph{Frequency-wise Energy Retention.}
To examine how individual spectral components are preserved across layers, we analyze the retained energy of each Laplacian frequency after $L$ propagation steps:
\begin{equation}
E(\lambda; L) = |\mu(\lambda)|^{L}.
\end{equation}
This quantity measures how strongly each frequency contributes to the representation after repeated message passing.

As shown in Fig.~\ref{fig:energy-cora} and Fig.~\ref{fig:energy-coauthorcs}, standard GCNs rapidly suppress intermediate and high-frequency components, consistent with classical oversmoothing behavior.
In contrast, Laplacian-LoRA preserves substantially more energy across the spectrum, particularly at moderate and larger eigenvalues.
This effect is especially pronounced on larger graphs such as CoauthorCS, where retaining higher-frequency components is crucial for maintaining discriminative representations.

These results show that Laplacian-LoRA learns a minimal and stable spectral correction that globally reduces contraction while preserving the low-pass inductive bias of GCNs.
Rather than eliminating oversmoothing, the method delays the depth at which spectral collapse occurs by slowing relative contraction and rebalancing frequency attenuation, in close agreement with our theoretical analysis.

\section{Conclusion}

We presented Laplacian-LoRA, a simple and interpretable spectral adaptation of GCNs that delays oversmoothing in deep graph networks. By introducing a low-rank, spectrally anchored correction, the method weakens per-layer contraction and shifts the depth at which representational collapse occurs, as confirmed by both performance trends and spectral diagnostics. While oversmoothing is delayed rather than eliminated and the reliance on partial Laplacian eigendecomposition may limit scalability to very large graphs, our results highlight oversmoothing as a depth-dependent transition that can be meaningfully modulated through principled spectral design. Future work may explore scalable spectral approximations, layer-dependent adaptations, or integration with complementary architectural mechanisms. Our implementation can be found here 
\href{https://github.com/Vamsi995/gcn-smooth}{\color{blue}https://github.com/Vamsi995/gcn-smooth}.

\end{document}